\documentclass[10pt,twocolumn,letterpaper]{article}

\usepackage{cvpr}      

\usepackage{graphicx}
\usepackage{amsmath}
\usepackage{amssymb}
\usepackage{booktabs}
\usepackage{multirow}

\usepackage[pagebackref,breaklinks,colorlinks]{hyperref}

\usepackage[capitalize]{cleveref}
\crefname{section}{Sec.}{Secs.}
\Crefname{section}{Section}{Sections}
\Crefname{table}{Table}{Tables}
\crefname{table}{Tab.}{Tabs.}

\def\cvprPaperID{*****} 
\def\confName{CVPR}
\def\confYear{2024}

\begin{document}

\title{Regularized PolyKervNets: Optimizing Expressiveness and Efficiency for Private Inference in Deep Neural Networks}

\author{Toluwani Aremu\\
Mohamed Bin Zayed University of Artificial Intelligence, UAE\\
{\tt\small toluwani.aremu@mbzuai.ac.ae}
}
\maketitle

\begin{abstract}
   Private computation of nonlinear functions, such as Rectified Linear Units (ReLUs) and max-pooling operations, in deep neural networks (DNNs) poses significant challenges in terms of storage, bandwidth, and time consumption. To address these challenges, there has been a growing interest in utilizing privacy-preserving techniques that leverage polynomial activation functions and kernelized convolutions as alternatives to traditional ReLUs. However, these alternative approaches often suffer from a trade-off between achieving faster private inference (PI) and sacrificing model accuracy. In particular, when applied to much deeper networks, these methods encounter training instabilities, leading to issues like exploding gradients (resulting in NaNs) or suboptimal approximations. In this study, we focus on PolyKervNets, a technique known for offering improved dynamic approximations in smaller networks but still facing instabilities in larger and more complex networks. Our primary objective is to empirically explore optimization-based training recipes to enhance the performance of PolyKervNets in larger networks. By doing so, we aim to potentially eliminate the need for traditional nonlinear activation functions, thereby advancing the state-of-the-art in privacy-preserving deep neural network architectures. Code can be found on GitHub at: \url{https://github.com/tolusophy/PolyKervNets/}
\end{abstract}

\section{Introduction}
\label{sec:intro}

The widespread adoption of Machine Learning as a Service (MLaaS) has encountered a substantial challenge in the form of privacy concerns associated with sensitive user data. To address these concerns, various privacy-preserving techniques have been developed and deployed. One promising approach seeks to perform inference directly on encrypted data \cite{cryptonet, delphi, crypten, cryptflow, secureml, xonn, cheetah, gazelle, minionn, mpcformer}, employing methods such as homomorphic encryption (HE) \cite{he} or multiparty computation (MPC) \cite{mpc}, including secret sharing (SS). In typical privacy-preserving inference (PI) protocols, HE/SS handles linear operations, while garbled circuits (GC) are used for nonlinear operations.

While these protocols efficiently handle linear operations, nonlinear operations present a significant challenge, often dominating runtime and incurring substantial costs in terms of storage and latency. To address this issue, researchers have explored different strategies, including budgeting \cite{cryptonas, deepreduce, deepreshape, autorep} for nonlinear functions (although this remains inefficient) and leveraging polynomial activation functions \cite{ramy, aespa, compact, FastAP, sisyphus, cryptinfer, vikas, capride} and polynomial kernel convolutions \cite{polykervnet}. The latter approach combines convolutions and polynomial activations, eliminating the need for separate nonlinear activation functions. According to \cite{delphi}, the complete replacement of ReLUs with polynomial activation functions i.e., $x^2$ (Quad) \cite{cryptonet} can result in impressive latency improvements of up to $3000\times$ and communication efficiency gains of up to $300\times$.

However, while these innovations have contributed to the design of privacy-friendly networks, they can introduce performance challenges, especially when fully replacing ReLUs \cite{sisyphus}. Previous studies \cite{cryptinfer, cryptonet, minionn, ramy} that have completely replaced ReLUs with low-degree polynomials have demonstrated improvements on smaller and simpler datasets and models. However, when these approaches are applied to larger datasets and more complex models, they encounter training instabilities, resulting in issues such as NaNs or significant performance degradation \cite{sisyphus}.

In this paper, we aim to solve these issues by finding the best training optimization-based training recipes for deep polynomial residual models, with a specific focus on PolyKervNets (PKNs) \cite{polykervnet}, a recent state-of-the-art technique which changes the convolution kernel from a linear to a polynomial one. PKNs leverage expressive polynomial convolution kernels (polykerv) to design dynamic functions that adapt at each layer. However, like other approaches, they face limitations, with the authors unable to extend them beyond 18 layers of residual networks. By getting the best training recipes for PKNs, we believe they can be extended to other polynomial-based approaches, as well as more recent networks i.e., ViTs. Our main contributions in this work are as follows:
\begin{itemize}
    \item We empirically demonstrate that using PKNs to redesign deeper networks leads to training instabilities, leading to gradient explosion, even in residual networks.
    \item We also introduce Regularized PolyKervNets (RPKNs), which utilizes a learnable regularization parameter that is capable of minimizing the occurrence of vanishing or exploding gradients, thereby enabling PKNs for much deeper residual networks. We also introduce Regularized PolyKervNets Activation (React-PKNs), which is basically R-PKNs but as an activation function and also gives more control over the choice of initialization. Our approach improved the depth limit of PKNs from 18 residual layers to 50.
    \item We also assess the performance of RPKNs in a knowledge distillation setting. We train a student RPKN model alongside its vanilla architecture as the teacher. Notably, both architectures are trained concurrently so that the student model can follow the teacher's learning journey. This improved the training stability as well as performance of RPKNs.
\end{itemize}


\section{Background}
\subsection{PolyKervNets}
Conventional CNN models predominantly rely on activation layers to introduce point-wise non-linearity into the machine learning model. In the work by \cite{polykervnet}, it was demonstrated that CNN architectures could achieve greater expressiveness by replacing the linear convolution layer with patch-wise non-linearity using the kernel trick, as described in \cite{kerv}. This innovative approach, known as polynomial kernel convolution (PolyKervNets or PKNs), and has been leveraged in several works \cite{AdderNet,KernelFI,Aremu2023}, expanded upon the ideas presented in \cite{kerv}. While \cite{kerv} primarily focused on fundamental computer vision tasks, \cite{polykervnet} explored the intersection between the requirements of privacy-preserving inference (PI) protocols and the core concepts from \cite{kerv}. They pushed the envelope by completely eliminating nonlinear layers, resulting in the creation of privacy-friendly variants of conventional CNN architectures.

We begin with the conventional convolution operator, which takes an input $x\in\mathbb{R}^n$ and weights $w\in\mathbb{R}^n$, and produces the following output:
\[
k_c(x,w)=x^T w+b
\]
where $b$ represents the bias term. In the context of PolyKervNets, a patch-wise nonlinear kernel is applied, defined as follows:
\[
k_p(x,w) = (x^T w+c_p)^{d_p}+b
\]
Here, $d_p$ ($d_p\in\mathbb{Z}^+$) denotes the polynomial degree, and $c_p$ ($c_p\in\mathbb{R}^+$) is a learnable balance factor. In essence, the polynomial degree extends the feature space to $d_p$ dimensions, and the balancing factor adjusts the non-linear terms. Consequently, the polynomial kernel not only captures linear relationships similar to the standard convolutional operator but also generates non-linear terms. These non-linear terms can be seen as an approximation of the output from an activation layer, making the activation learnable and enhancing its expressiveness with each use.

\subsection{Problems with PKNs}
While polynomial approximations of nonlinear functions are the preferred approach for designing PI-friendly networks, they either suffer from training instabilities as the model gets deeper or a high accuracy deficit compared to their vanilla variants. Though PKNs provide an active way of co-designing polynomial kernels suitable for each convolution layer, it also terribly suffers from the same issues as the depth increases. In fact, while the method was able to achieve proximal results with respect to the vanilla architectures, training stability decreased as depth increased. It was a struggle to train networks larger than PKR-4 (ResNet-18) as they either returned NaN loss values, or diverge to a local minimal of an accuracy around 10\%. In this section, we will show the problems using three experiments;

\begin{table}[htbp]
\centering
\caption{MSE Loss between Outputs of PKN Variants and their Vanilla Counterparts}
\label{mse}
\resizebox{\linewidth}{!}{%
\begin{tabular}{|l|c|c|c|}
\hline
Method & PKN ($d_p=2, c_p=0$) & PKN ($d_p=3, c_p=0$) & PKN ($d_p=4, c_p=0$) \\
\hline
FC only & 0.1177 & 0.111 & 0.1125 \\
CNN3 & 2.1792 & 9.7036 & 85.2739 \\
Lenet & 2.3994 & 55.0785 & NaN \\
VGG11 & 0.0092 & 0.0092 & 0.0092 \\
VGG16 & 0.0023 & 0.0023 & NaN \\
Resnet18 & NaN & NaN & 0.6143 \\
Resnet32 & NaN & 0.8051 & NaN \\
Resnet50 & NaN & NaN & NaN \\
\hline
\end{tabular}
}
\end{table}

\noindent \textbf{Mean Squared Error (MSE) Loss of PKNs Compared to Their Vanilla Variants}: For this experiment, we employed multiple networks and initialized their weights randomly. Subsequently, we cloned these networks and replaced the convolution layers with \textit{polykerv} layers, removed the ReLU layers, and replaced the max-pooling layers with average pooling layers. These models were frozen, and without having to train, we conducted a forward pass using the same randomly generated input for each network. The MSE Loss for the final outputs of each PKN variant was then computed. In Table \ref{mse}, we can see the inconsistencies with regards to each network. Most of the deeper networks returned NaN values, which is caused by the forward activation values growing exponentially, leading to an explosion. \\

\begin{table}[htbp]
\centering
\caption{Train Loss and Test Accuracy on CIFAR-10 (Training from Scratch)}
\label{scratch}
\resizebox{\linewidth}{!}{%
\begin{tabular}{|l|c|c|c|c|}
\hline
Method & ReLU & PKN ($d_p=2$) & PKN ($d_p=3$) & PKN ($d_p=4$) \\
\hline
CNN3 & 1.3774 (53.04) & 1.4432 (52.19) & 1.4526 (51.36) & 1.6928 (43.07) \\
Lenet & 1.3312 (53.67) & 1.4977 (49.13) & 1.7463 (39.30) & NaN \\
VGG11 & 0.5133 (73.86) & 0.5912 (71.43) & NaN & NaN \\
VGG16 & 0.4848 (73.68) & 0.5958 (71.07) & NaN & NaN \\
Resnet18 & 0.8198 (66.60) & 0.8336 (66.38) & NaN & NaN \\
Resnet32 & 0.8953 (63.55) & NaN & NaN & NaN \\
Resnet50 & 1.4356 (47.22) & NaN & NaN & NaN \\
\hline
\end{tabular}
}
\end{table}

\noindent \textbf{Standard Train-from-Scratch Recipe}: For this experiment, we followed the same steps as in the first experiment. This time, we trained all models from scratch on the CIFAR-10 dataset. Note that, except for CNN3 and LeNet, the other networks were obtained from PyTorch's library. These networks were originally designed for $224\times224$ images, but we trained them on $32\times32$ images for 100 epochs using the SGD with momentum algorithm (with a learning rate of 0.001 and momentum of 0.5). The purpose of this experiment was to compare the performance of each PKN variant with that of ReLU-based networks and not necessarily achieve optimal results for these networks. We initialized $c_p$ as $0.5$ for each degree of PKN, and the results showed comparable performance for smaller PKN networks with $d_p=2$ (see Table \ref{scratch}). Unfortunately, this comparison was limited to ResNet-18, as larger models returned NaN values. For other degrees, the performances were less promising, suggesting a faster explosion in gradients as $d_p$ increased.

\begin{table}[htbp]
\centering
\caption{Train Loss and Test Accuracy on CIFAR-10 (Finetuning)}
\label{pretrain}
\resizebox{\linewidth}{!}{%
\begin{tabular}{|l|c|c|c|c|}
\hline
Method & ReLU & PKN ($d_p=2$) & PKN ($d_p=3$) & PKN ($d_p=4$) \\
\hline
Resnet18 & 0.0613 (80.84) & 0.6479 (79.55) & NaN & NaN \\
Resnet32 & 0.0402 (82.11) & NaN & NaN & NaN \\
Resnet50 & 0.0216 (84.30) & NaN & NaN & NaN \\
\hline
\end{tabular}
}
\end{table}

\noindent \textbf{Fine-tuning Recipe}: For this experiment, we followed the same steps as in the second experiment. This time, we leveraged pre-trained models obtained from PyTorch's library and fine-tuned each network, including the PKN variants, using the SGD with momentum algorithm (with a learning rate of 0.001 and momentum of 0.9). While these networks were originally designed for $224\times224$ images, we acknowledge that achieving optimal results on $32\times32$ images may be challenging. Again, our primary objective is to highlight the issues with PKNs. We initialized $c_p$ as $0.5$ for each degree of PKN, and the results presented here also reinforce our point about the limitations of PKNs in terms of depth (see Table \ref{pretrain}).

\section{Regularized PKNs (R-PKNs)}

In order to address the high sensitivity issue observed in PKNs, we introduce a learnable parameter $a_p$ ($a_p\in\mathbb{R}^+$) that modifies the standard PKN kernel as follows:

\[
k_p(x,w) = a_p((x^T w+c_p)^{d_p}+b)
\]

Here, $a_p$ serves as a regularizing factor for the weights and coefficients associated with each input $x$ across all layers $f(x)$ within the learner $\theta$ without having to use the gradient clipping method. We refer to this enhanced variant as Regularized PKNs, or R-PKNs for short.

However, if we expand and separate the R-PKN kernel $k_p(x,w)$ as follows:

\[
k_p(x,w) = \sum_{i=0}^{d_p}{a_p(c_{p}^{d_p -i}(x^T w)^{i})}
\]

We can create a learnable low-degree polynomial activation function, specifically for $d_p=2$, which provides greater control and flexibility. The new form of the activation function is expressed as:

\[
f(x) = a_p(x^2) + b_p(x) + c_p
\]

In this expression, $x$ represents the output of a convolutional kernel $k_c(x,w) = x^T w$, $b_p$ is a learnable function that can be flexibly initialized and corresponds to $2a_pc_p$ in the expanded function, and $c_p$ is defined as $c_p=c_p^2$. This modification simplifies the deployment of R-PKNs in pre-trained networks. We refer to this version as Regularized PolyKervNet Activation, or React-PKNs for short, which we use extensively in our experiments.

\section{Empirical Journey}

Our objective is to discover the optimal training strategies to enhance the performance of polynomial activation functions within very deep networks, with a specific focus on PKNs. To achieve this goal, we have organized our experiments into distinct phases. We center our attention on the ResNet architecture for our experimental investigations and illustrate the effectiveness of Regularized PKNs (R-PKNs) on the CIFAR-10 dataset.

In \cite{polykervnet}, it was observed that PKNs can produce results closely aligned with their vanilla counterparts but tend to encounter instability as the network depth increases. In fact, when adapting PKNs to ResNet-32 and ResNet-50 architectures, it returned NaN. In response to this challenge, we embark on a journey to identify the most effective optimization-based strategies for training these networks.

To augment the CIFAR-10 dataset, we introduce horizontal image flipping with a 50\% probability and apply the \textit{colorjitter} transformation. In an effort to maximize the potential of models available in the PyTorch library, we upscale the image size from $32\times32$ to $224\times224$ pixels. The models we employ are pre-trained, and we fine-tune all layers on the CIFAR-10 dataset. Our default batch size is 128.

In the following sections, we will delve into the intricacies of each experiment, providing comprehensive details, discussing the outcomes, and drawing conclusions \textit{based on the insights gained from our experiments}.

\subsection{Impact of Learning Rate}
We begin with the optimal results obtained from each of the vanilla ResNet models we utilized. We trained each model using both the SGD and Adam optimizers (Table \ref{tab:accuracy_comparison}) with a learning rate of $3 \times 10^{-4}$ for 50 epochs while employing the ReduceLRonPlateau scheduler. We limited the training to 50 epochs because the optimal results were already achieved by the 15th epoch.

\begin{table}[htbp]
\centering
\caption{Baseline Accuracy Comparison of ReLU-Based ResNet Models on CIFAR-10 Using Adam and SGD Optimizers (\%). These results establish the foundational performance benchmarks for our subsequent comparative analysis.}
\label{tab:accuracy_comparison}
\resizebox{\linewidth}{!}{%
\begin{tabular}{|l|c|c|c|c|c|}
\hline
Method & ResNet-10 & ResNet-14 & ResNet-18 & ResNet-32 & ResNet-50 \\
\hline
SGD & 87.88 & 89.24 & 94.35 & 94.77 & 95.33 \\
ADAM & 91.64 & 93.21 & 94.32 & 94.72 & 95.16 \\
\hline
\end{tabular}
}
\end{table}

The table above serves as the baseline for every training approach we use with respect to the R-PKN-based ResNet models. To simplify the model names and reduce confusion, we adopt a naming convention similar to that of PKNs. Specifically, ResNet-10 becomes RPKR-10, ResNet-14 becomes RPKR-14, ResNet-18 becomes RPKR-18, ResNet-32 becomes RPKR-32, and ResNet-50 becomes RPKR-50, where RPKR stands for Regularized PolyKervResNet. Also, since we are leveraging the React-PKNs, our learnable parameters' initialization are: $a_p=0.009, b_p=0.5, c_p=0.47$. These were gotten from a study \cite{vikas} on Polynomial Activation Functions. Finally, we trained these variants for 200 epochs to ensure that we achieve the best results.

\begin{table}[htbp]
\centering
\caption{Evaluating the Effects of Learning Rates on RPKR Variants Performance in CIFAR-10 Dataset. This comparison with the baseline performances (referenced in Table \ref{tab:accuracy_comparison}) reveals that smaller RPKR networks employing a learning rate of SGD 3e-4 demonstrate performances closely matching the baseline. Conversely, the larger RPKR networks underperform in comparison, indicating a potential scalability issue with increased network size under this learning rate regime.}
\label{tab:rpkr_accuracy}
\resizebox{\linewidth}{!}{%
\begin{tabular}{|l|c|c|c|c|c|}
\hline
Method & RPKR-10 & RPKR-14 & RPKR-18 & RPKR-32 & RPKR-50 \\
\hline
SGD (lr=1e-3) & 85.64 & 82.65 & 82.78 & 67.43 & 68.17 \\
SGD (lr=3e-4) & 86.3 & 85.9 & 87.08 & 85.41 & 66.54 \\ \hline
ADAM (lr=1e-3) & 83.36 & 82.06 & 82.61 & 37.35 & 76.15 \\
ADAM (lr=3e-4) & 85.72 & 84.82 & 87.09 & 85.61 & 58.3 \\
\hline
\end{tabular}
}
\end{table}

We then train RPKRs with different learning rates (1e-3 and 3e-4) without using schedulers and report their results in Table \ref{tab:rpkr_accuracy}. We observe that using a smaller learning rate for RPKRs generally leads to more stable results, even though they may not closely match the results of the vanilla networks in Table \ref{tab:accuracy_comparison}. However, thanks to our learnable parameters, we can now extend the depth limit from ResNet-18 (PKR-4/RPKR-18) to ResNet-32 (RPKR-32). In the PKN paper, as well as in our preliminary studies in the earlier sections of this paper, it was noted that training a ResNet-32 network with PKNs was not feasible. This represents a significant improvement, although RPKR-50 is still not fully stable. In summary, for RPKNs, a smaller learning rate tends to yield better results.

\subsection{Impact of Batch Sizes}
Using a learning rate of $3\times10^{-4}$ for its stability, as observed in our earlier experiments, we investigated the impact of batch size on RPKNs. In contrast to our previous experiments where schedulers were not employed, we utilized the ReduceLROnPlateau scheduler for this specific set of experiments.

\begin{table}[htbp]
\centering
\caption{Comparative Performance of RPKR Variants with Varied Optimization Methods and Batch Sizes. Employing the ReduceLROnPlateau scheduler, we trained various RPKR models under different batch sizes. Our findings indicate that the Adam optimizer consistently outperforms SGD in PKN methods. Notably, the choice of batch size plays a crucial role in reaching optimal solutions. Smaller networks align closely with baseline performance at batch sizes of 1, 4, and 128, whereas a larger batch size of 512 leads to suboptimal results.}
\label{tab:rpkr_performance}
\resizebox{\linewidth}{!}{%
\begin{tabular}{|l|c|c|c|c|c|}
\hline
Method & RPKR-10 & RPKR-14 & RPKR-18 & RPKR-32 & RPKR-50 \\
\hline
SGD (bs=128) & 75.71 & 68.64 & 88.77 & 84.39 & 70.56 \\
SGD (bs=512) & 48.67 & 58.16 & 78.23 & 62.86 & 70.73 \\ \hline
ADAM (bs=128) & 86.15 & 87.16 & 88.28 & 86.15 & 74.2 \\
ADAM (bs=512) & 81.23 & 84.55 & 86.31 & 83.12 & 68.43 \\
ADAM (bs=4) & 88.79 & 88.35 & 86.75 & 77.61 & 75.65 \\
ADAM (bs=1) & 89.01 & 89.98 & 87.21 & 79.31 & 75.07 \\
\hline
\end{tabular}
}
\end{table}

In Table \ref{tab:rpkr_performance}, increasing the batch size did not lead to an improvement in the performance of RPKRs. In fact, for smaller RPKRs, there was a significant increase in accuracy when we reduced the batch size from 128 to 4 and 1. However, training networks using a very small batch size requires a considerable amount of time, which may not be practically efficient. Additionally, as the model size increased, the performance with respect to batch size decreased.

While SGD struggled to consistently deliver satisfactory results, employing the Adam optimizer with a batch size of 128 and a scheduler consistently outperformed Adam without a scheduler. This brought us closer to our objective of achieving optimal performance, particularly for larger model sizes. Notably, in \cite{polykervnet}, PKRs achieved significantly improved results when using Adam with a learning rate of 1e-3 and a step scheduler that reduces the step size by a factor of 0.1 every 80 epochs. We hypothesize that this difference in performance could be attributed to the choice of hyperparameters and variations in training configurations. Our primary objective is to identify an optimal training configuration or recipe that can effectively extend the applicability of PKNs and potentially other polynomial-based neural network architectures to models with substantial depth. Our aim is to achieve this extension without resorting to gradient-constraining techniques, such as gradient clipping.

\subsection{Training with MoMo}
During the course of our study, we encountered an optimization algorithm that automatically learns and adapts the step sizes for momentum-based optimizers. MoMo \cite{momo} utilizes momentum estimates of the batch losses and gradients sampled at each iteration to construct a model of the loss function. It then approximately minimizes this model at each iteration to compute the next step. This allows us to train RPKNs without the need for using schedulers.

\begin{table}[htbp]
\centering
\caption{Comparative Efficiency of RPKR Models with MoMoAdam Optimizer: RPKR models using MoMoAdam outperformed those using Adam and scheduler-based methods, especially in the RPKR-32 model. The study highlights the efficacy of very small learning rates for optimal performance in deeper architectures, as evidenced by the results at the 200th epoch, underscoring the intricacies of training complex networks.}

\label{tab:rpkr_results}
\resizebox{\linewidth}{!}{%
\begin{tabular}{|l|c|c|c|c|c|}
\hline
Method & RPKR-10 & RPKR-14 & RPKR-18 & RPKR-32 & RPKR-50 \\
\hline
MoMoAdam (lr=1e-2) & 86.08 & 78.97 & 82.69 & 42.48 & 55.49 \\
MoMoAdam (lr=1e-3) & 87.73 & 86.38 & 88.12 & 83.7 & 75.21 \\
MoMoAdam (lr=3e-4) & 87.07 & 88.48 & 89.65 & 88.37 & 72.16 \\
MoMoAdam (lr=3e-5) & 86.48 & 78.07 & 82.35 & 82.62 & 83.08 \\
MoMoAdam (lr=5e-6) & 70.06 & 77.91 & 82.1 & 82.44 & 83.38 \\
\hline
\end{tabular}
}
\end{table}

Note that we aim to achieve similar or close results to those shown in Table \ref{tab:accuracy_comparison}. To investigate the impact of MoMoAdam (a version of MoMo adapted for the Adam optimizer) on RPKNs, we conducted experiments with various initial learning rates. The results in Table \ref{tab:rpkr_results} indicate that utilizing MoMoAdam yields significantly better results compared to using just Adam or Adam with schedulers.

When comparing the performance of RPKRs and vanilla ResNets using a learning rate of 3e-4, we observe an increase in performance when employing MoMoAdam instead of Adam. However, achieving satisfactory results with ResNet-50 remains a challenge. Notably, we discovered that using a smaller learning rate than 3e-4 yielded decent performance in RPKR-50, albeit requiring longer training times. This suggests that larger models using R-PKNs benefit from smaller learning rates but require extended training periods. Our observations indicate that training with learning rates of 3e-5 and 5e-6 showed potential for improvement, with slower training being attributed to the initial learning rate.

\subsection{Impact of Knowledge Distillation}
In this section, we explore the application of knowledge distillation (KD), a well-established technique in deep learning. In traditional KD, a trained teacher network ($\theta_t$) provides guidance for training a smaller student network ($\theta_s$). The primary objective is to transfer the knowledge and generalization capabilities of the teacher to the student, allowing the student to achieve performance close to that of the teacher.

In our specific context, we employ KD to delve deeper into the potential advantages of Regularized PolyKervNet Activation (React-PKNs). Our approach involves training a student network alongside its vanilla counterpart, which acts as the teacher. This novel method entails concurrent training of both architectures, enabling the student model to closely follow the learning trajectory of the teacher. By incorporating KD with React-PKNs, our aim is to enhance training stability and overall performance across all residual-based PKN variants we have adapted.

The KD process involves forwarding the input $x$ through the teacher network $\theta_t$, calculating the cross-entropy loss, and performing backpropagation. Subsequently, the output generated by the updated teacher's weights serves as the label for the student network. The Kullback-Leibler (KL) divergence loss between the student and teacher is then used to update the student model.

\begin{table}[htbp]
\centering
\caption{Accuracy (\%) of RPKR Models Trained Using Distilled Knowledge from their Vanilla Variants. This comparison indicates that implementing knowledge distillation in PKNs closely aligns the student models' performance with that of their teacher models.}
\label{tab:rpkr_accuracy2}
\resizebox{\linewidth}{!}{%
\begin{tabular}{|l|c|c|c|c|c|}
\hline
Method & RPKR-10 & RPKR-14 & RPKR-18 & RPKR-32 & RPKR-50 \\
\hline
ADAM (lr=3e-4) & 90.00 & 91.68 & 92.11 & - & - \\
MoMoAdam (lr=3e-4) & 91.61 & 92.87 & 94.00 & - & - \\
MoMoAdam (lr=3e-5) & - & - & - & 87.02 & 88.71 \\
MoMoAdam (lr=5e-6) & - & - & - & 86.77 & 88.06 \\
\hline
\end{tabular}
}
\end{table}

Our experiments, as shown in Table \ref{tab:rpkr_accuracy2}, reveal that employing a learning rate of 3e-4 in Adam and MoMoAdam for both teacher and student models leads to improved results and significantly reduces accuracy discrepancies, especially for smaller architectures (10, 14, 18). However, maintaining stability for larger models becomes challenging with this learning rate setting. As a remedy, we reduced the learning rate to 3e-5 and 5e-6 for RPKR-32 and RPKR-50, while keeping the teacher variants (ReLU-based) at 3e-4. For smaller networks, we did not perform further experiments at this stage. Our results indicate notable improvements for these larger networks, thus addressing the stability concerns. Better results can be achieved than those presented here with longer training.

\section{Results Discussion and Future Work}

In this section, we summarize the key findings from our experiments and outline potential directions for future research:

\subsection{Key Findings}

\begin{itemize}
    
    \item In our experiments, we observed that employing smaller learning rates generally resulted in more stable outcomes for RPKRs, particularly for larger network architectures. Although RPKRs did not precisely match the performance of vanilla networks, using smaller learning rates significantly improved stability. This allowed us to extend the depth limit to ResNet-32, with RPKR-50 showing promising progress. In summary, for RPKNs, smaller learning rates tend to yield more favorable outcomes.

    \item The impact of increasing batch sizes on RPKR performance was inconsistent. Surprisingly, reducing the batch size from 128 to 4 and even 1 led to substantial accuracy gains for smaller RPKRs. However, extremely small batch sizes may not be practical due to prolonged training times. Additionally, as model size increased, the relationship between batch size and performance weakened. Utilizing the Adam optimizer with a scheduler consistently outperformed Adam without a scheduler, particularly for larger models.

    \item Our experiments with MoMoAdam, an optimizer that adjusts step sizes using momentum estimates, demonstrated significant performance improvements for RPKRs. Smaller learning rates, such as 3e-5 and 5e-6, produced respectable results for larger RPKRs, albeit with extended training durations. This suggests that larger RPKR models benefit from smaller learning rates but require longer training periods.

    \item Knowledge distillation (KD) proved effective in enhancing the performance and stability of RPKRs. The concurrent training of student and teacher models using React-PKNs and the Kullback-Leibler (KL) divergence loss resulted in substantial accuracy improvements. Lower learning rates, such as 3e-5 and 5e-6, demonstrated potential for further enhancement, especially for larger RPKR models. KD addressed stability concerns and brought RPKR models closer to achieving optimal performance.
\end{itemize}

\subsection{Future Directions}

In conclusion, while we have made significant progress in our quest to optimize the training of polynomial-based networks, particularly RPKNs, it's important to acknowledge the constraints that have shaped the scope of our study. Due to limitations in time and resources, as this project was conducted as part of a course, we were unable to conduct additional experiments that would explore the full extent and boundaries of the methods under investigation.

Nonetheless, we firmly believe that our empirical journey represents a valuable contribution to the understanding and advancement of these models. Our findings, although not exhaustive, offer important insights into training strategies for polynomial-based networks. As we were unable to push these methods to their limits, we outline several intriguing avenues for future research:

\begin{itemize}
    \item Investigating the potential benefits of combining R-PKNs with gradient clipping to determine if this approach can yield comparable or superior results in terms of stability and overall performance.

    \item Exploring layer-wise learning rate initialization, where deeper layers are assigned different learning rates than initial layers, in order to further optimize the training process for polynomial-based networks. A quick experiment with this gave RPKR-50 an accuracy of 87.9\% without requiring tuning or knowledge distillation.

    \item Exploring alternative optimization techniques, such as Quasi-Newton based approaches, to determine if certain types of optimizers exhibit superior performance and convergence properties when applied to polynomial-based networks.

    \item Extending the scope of our conclusions to assess whether they are applicable to other polynomial-based approaches, beyond RPKRs, in various deep learning scenarios.

    \item Evaluating the generalizability of our approach to different datasets and model architectures, such as Vision Transformers (ViTs), to determine its effectiveness in a broader context.
\end{itemize}

{\small
\bibliographystyle{ieeetr}
\bibliography{egbib}
}

\end{document}